\documentclass{article}

    \usepackage[preprint, nonatbib]{neurips_2022}

\usepackage[utf8]{inputenc} 
\usepackage[T1]{fontenc}    
\usepackage{hyperref}       

\usepackage{url}            
\usepackage{booktabs}       
\usepackage{amsfonts}       
\usepackage{nicefrac}       
\usepackage{microtype}      
\usepackage{xcolor}         

\usepackage{graphicx}
\usepackage{amsmath}
\usepackage{amssymb}
\usepackage{mathtools}

\usepackage{algorithm}
\usepackage{algpseudocodex}
\algnewcommand{\LeftComment}[1]{\Statex \(\triangleright\) #1}

\usepackage{biblatex}
\usepackage{cleveref}
\graphicspath{figures}

\title{Recursive Neural Programs: Variational Learning of Image Grammars and Part-Whole Hierarchies}

\author{Ares Fisher, Rajesh P. N. Rao\\
Paul G. Allen School of Computer Science and Engineering\\
University of Washington, Seattle\\
{\tt\small \{aresf, rao\}@cs.washington.edu} 
}

\begin{document}

\maketitle

\begin{abstract}
    Human vision involves parsing and representing objects
    and scenes using structured representations based on part-whole hierarchies. Computer vision and machine learning researchers have recently sought to emulate this capability using capsule networks, reference frames and active predictive coding, but a generative model formulation has been lacking. We introduce Recursive Neural Programs (RNPs), which, to our knowledge, is the first neural generative model to address the part-whole hierarchy learning problem. RNPs model images as hierarchical trees of probabilistic sensory-motor programs that recursively reuse learned sensory-motor primitives to model an image within different reference frames, forming recursive image grammars. We express RNPs as structured variational autoencoders (sVAEs) for inference and sampling, and demonstrate parts-based parsing, sampling and one-shot transfer learning for MNIST, Omniglot and Fashion-MNIST datasets, demonstrating the model's expressive power. Our results show that RNPs provide an intuitive and explainable way of composing objects and scenes, allowing rich compositionality and intuitive interpretations of objects in terms of part-whole hierarchies.

\end{abstract}

\section{Introduction}
Human visual cognition relies heavily on hierarchical relationships between objects and their parts. For example, a human face can be modeled as a hierarchical tree of parts, each part's relative position specified within a local reference frame: eyes, nose, mouth etc.\ positioned within the face's reference frame, the parts of an eye (eyebrow, eyelid, iris, pupil etc.) positioned within the eye's reference frame, and so on. To emulate such a capability, a computer vision system needs to not only learn what a part looks like (shapes, contours, colors etc. as in current deep convolutional networks) but also the relative transformation of the part within a local reference frame, and do this recursively in order to compose a human face (or a Picasso painting). 

Beyond vision, nested structure and hierarchical parts-based decompositions are ubiquitous in human attributes such as natural language (texts, chapters, paragraphs, sentences, words, characters) and complex behaviors (cooking a recipe, driving to work, etc.). Such recursive modeling confers the important property of compositionality \cite{lake_human-level_2015}: the same building blocks can be hierarchically and recursively composed into an endless variety of possible patterns, allowing an agent to "imagine" novel configurations of parts (e.g., for creating new solutions to problems), and recognize new configurations of known parts for zero-shot generalization. The challenge lies in learning a model of the parts and their transformations that is recursive and composable. Existing approaches for parsing tree-structured data \cite{eslami_et_al_attend_2016, lake_human-level_2015, hinton_matrix_2018, hinton_how_2021, mnih_et_al_recurrent_2014, socher_et_al_parsing_2011} are either not recursive \cite{eslami_et_al_attend_2016, mnih_et_al_recurrent_2014}, not compositional \cite{socher_et_al_parsing_2011}, not generative \cite{hinton_matrix_2018, hinton_how_2021}, or not differentiable \cite{lake_human-level_2015}. Indeed, the lack of a smooth ``program space" has been a challenge in this regard.

We introduce recursive neural programs (RNPs), which address this problem by creating a fully differentiable recursive tree representation of sensory-motor programs. Our model builds on past work on Active Predictive Coding Networks \cite{gklezakos_active_2022} in using state and action networks but is fully generative, recursive, and probabilistic, allowing a structured variational approach to inference and sampling of neural programs. The key differences between our approach and existing approaches are: 1) Our approach can be extended to arbitrary tree depth, creating a "grammar" for images that can be recursively applied 2) our approach provides a sensible way to perform gradient descent in hierarchical ``program space,'' and 3) our model can be made adaptive by letting information flow from children to parents in the tree, e.g., via prediction errors \cite{jiang_preston_l_dynamic_2021,gklezakos_active_2022}.

\section{Recursive Neural Programs}
\begin{figure*}[t]

    \includegraphics[width=\textwidth]{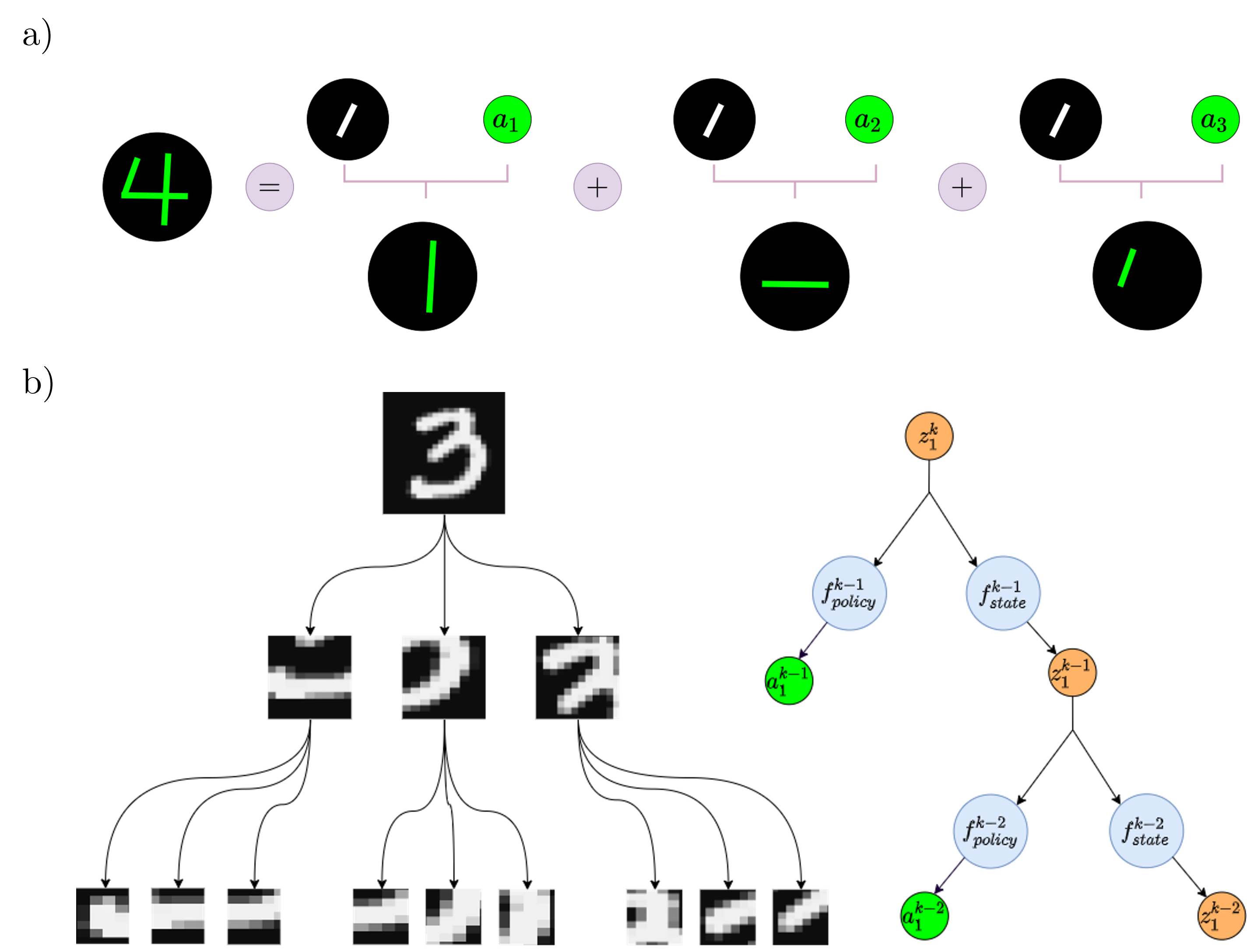}
    
    \caption{
    \textbf{Image Parsing as a Sequence of Transformed Primitives}. ({\bf a}) A $4$ can be constructed by generating three identical straight lines (black circles) and transforming them according to parameters $a$ to place them in the appropriate location. ({\bf b}) Left: An MNIST digit decomposed into an abstraction tree of parts, each of which is further decomposed into smaller sub-parts. Right: Schematic representation of a parsing tree produced by a recursive neural program. The digit is described as a ``program,'' represented by the vector $z^k$, which generates functions $f^k_{state}$ and $f^k_{policy}$ to construct the digit by generating parts and transforming them according to the action vector $a^k$ (position, scale, ...) within the digit's reference frame. Each part is in turn described by a program $z^{k-1}$, which generates smaller parts transformed according to $a^{k-1}$ within that part's reference frame.
}
    \label{fig-1}
\end{figure*}

We describe a 2-level Recursive Neural Program (RNP), though the architecture can be generalized to more levels. Consider the problem of parsing an image of a digit at two levels ($k=\{1,2\}$) of an abstraction tree (\cref{fig-1}), e.g., in terms of larger parts and smaller strokes (henceforth referred to as parts and sub-parts). A top-level program ($k=2$) generates the digit in terms of parts and a bottom-level program ($k=1$) generates each large part as a sequence of smaller parts and their transformations within the larger part's reference frame. Each program is expressed as an interaction between two recurrent functions, a state-transition function (or state-based forward model) that predicts the next state  $z^k_{t+1} = f^k_{state}(z^k_t, a^k_t) $, and an action transition function (policy) $a^k_{t+1} = f^k_{policy}(z^k_t, a^k_t)$ (\cref{fig-1}b, \cref{fig-2}, \cref{alg1}; in this paper, we assume actions correspond to transformations of parts). This is similar to the next-state and policy functions in a partially observable Markov decision process (POMDP \cite{kaelbling_planning_1998}). 

A program at tree depth $k$, represented by the state vector $z^k_t$, generates a fixed-length sequence of $\tau^k$ lower level states $ S^{k-1} = \{z^{k-1}_1, ..., z^{k-1}_{\tau^k}\}$ and their transformations $T^{k-1} = \{a^{k-1}_1, ..., a^{k-1}_{\tau^k}\}$. The state $z^k_t$ can be decoded into an image patch $\hat{x}^k_t$ that corresponds to a stroke or other image feature, then transformed according to $g(\hat{x}^k_t, a^k_t)$ to place it on a ``canvas'' (here $a$ refers to parameters of an affine transform on a grid, where $g$ is the bilinear interpolation function \cite{jaderberg_et_al_spatial_2015}.  The transformed images are added together at each time step, such that each step increasingly approximates the target image represented by $z^{k}$ (\cref{fig-2}b). This method allows us to reuse the same strokes with different transformations. For example, if $z^k_t$ represents a $\mathbf{4}$, $S^{k-1}$ can represent three straight lines, and $T^{k-1}$ are the transformations that orient and place them in the configuration of a $\mathbf{4}$ (\cref{fig-1}a). 

The above model can be made recursive, with generation performed in a depth-first manner: each $z^k_t$ generates the program for a sequence $\{ z^{k-1}_1,..., z^{k-1}_{\tau^{k-1}} \}$. $z^k_{t+1}$ begins after $z^k_t$ terminates. Here we use the decoded patches $\{\hat{x}^{k-1}_1, ..., \hat{x}^{k-1}_{\tau^{k-1}}\}$ as accumulated evidence to update $z^k_t$ (similar to other predictive coding models \cite{jiang_preston_l_dynamic_2021, gklezakos_active_2022}).

\subsection{Model architecture}

\begin{figure}
    \includegraphics[width=1\textwidth]{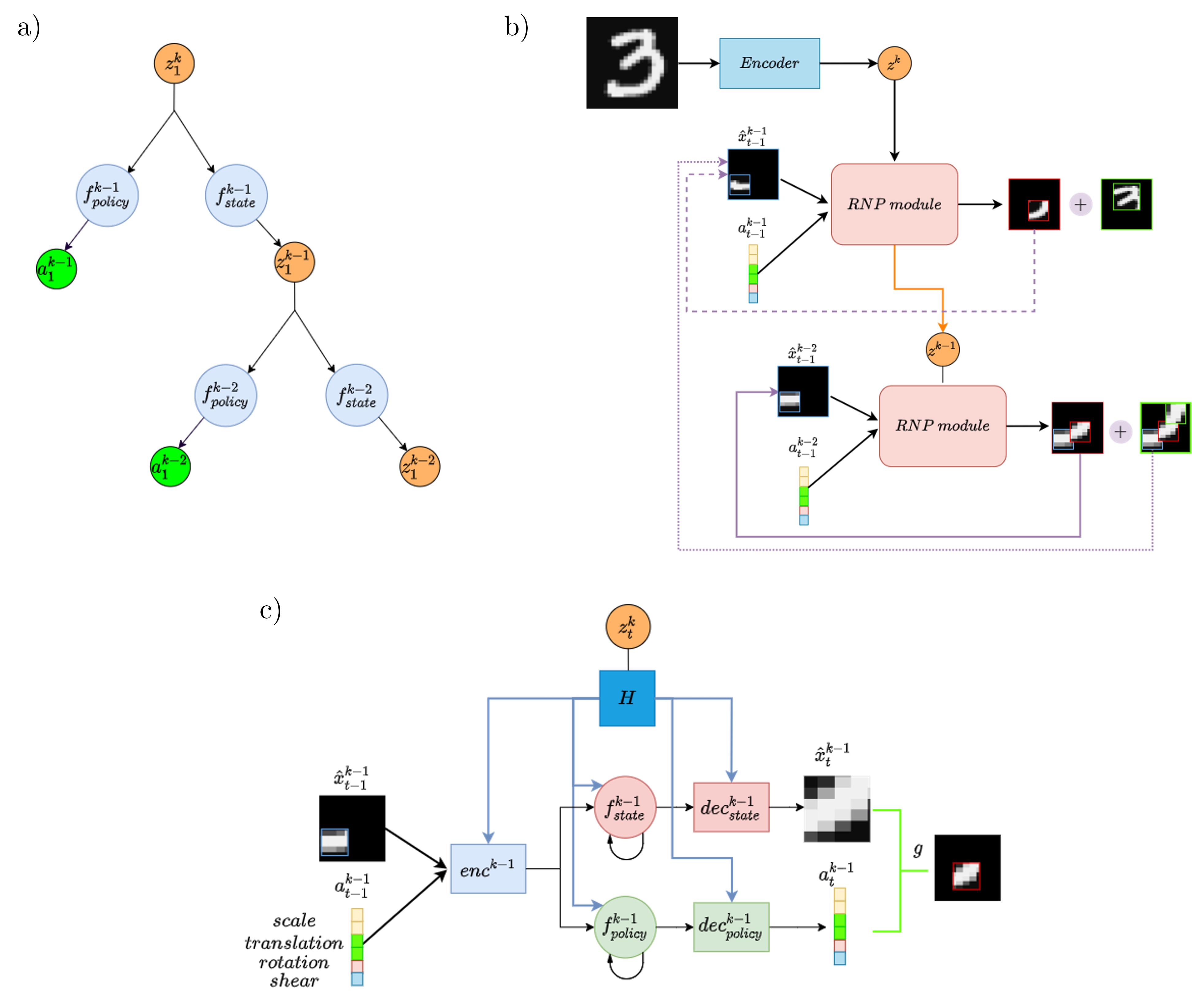}
    
    \caption{
    \textbf{Recursive Neural Program for Parsing Images}. ({\bf a}) Parsing tree for a digit, same as \cref{fig-1}b. ({\bf b)} Representation of parsing tree in ({\bf a}) with a neural network, where $z^k$ is generated by an encoder network. At time $t$, the network receives the most recent estimate of the part/sub-part and $a^k_{t-1}$ as input, and generates a prediction of the next transformed part or sub-part. Transformed parts and sub-parts are summed at their respective levels. Purple lines indicate recurrence. In our implementation, the top level $k$ receives the output of the program $k-1$ as input (dotted purple line) as opposed to its own output (dashed purple line). ({\bf c}) The RNP module consists of a hypernet $\mathcal{H}$, which parameterizes $f^k_{state}$, $f^k_{policy}$ and auxiliary networks to perform the computation shown in ({\bf b}).
}
    \label{fig-2}
\end{figure}

In a two-level RNP (\cref{fig-1}, \cref{fig-2}), the top-level program $z^2$ parameterizes two recurrent neural networks (RNNs) $f^2_{state}$ and $f^2_{policy}$ via a hypernetwork \cite{ha_hypernetworks_2016} (\cref{fig-2}b,c). The hypernet $\mathcal{H}$ is an MLP with seven heads, five of which generate parameters for the level-specific networks: an encoder (3-layer MLP), $\hat{e} = \mathcal{E}^k(\hat{x}^k_t, a^k_t)$, where $\hat{e}$ is the input to the $f^k_{state}$ and $f^k_{policy}$ networks; two recurrent networks $f^k_{state}$ and $f^k_{policy}$, with hidden size $| z | $, and their initial hidden states; and two decoders (3-layer MLP's): $\hat{x}^k_{t+1} = \mathcal{D}^k(z^k)$ generates an image patch from $z^k$, and $a^k_{t+1} = \mathcal{T}^k(h_{policy}(z^k_t, a^k_t)$ translates the hidden state of $f^k_{policy}$ into parameters $a^k_{t+1}$ (scaling, translation, rotation and shear) that transform $\hat{x}^k_{t+1}$. The remaining two heads provide initialization values $\hat{x}^k_0,\ a^k_0$ to initialize the sequence generation. More implementation and training details are in the Supplementary Material section.

We train the model described above by exploiting the end-to-end differentiability of the architecture, minimizing the reconstruction loss between all transformed sub-parts and the target image $\hat{x}$, regularized by the reconstruction at the level of parts:

\begin{align}
        \mathcal{L} = &\Vert \sum_{t_2 = 1}^{\tau^2} g( \sum_{t_1 = 1}^{\tau^1} g(\hat{x}^1_{t_1}, a^1_{t_1}), a^2_{t_2}) - x \Vert^2_2 + 
      \frac{1}{\tau^2} \Vert \sum_{t_2 = 1}^{\tau^2} g(\hat{x}^2 , a^2_{t_2}) - x^{patch}_{t_2} \Vert^2_2 
      \label{eq-1}
\end{align}

where $\tau^2$ and $\tau^1$ are the number of level-2 and level-1 time steps respectively, $x$ is the target image and $x^{patch}_{t_2}$ is the image patch generated by transforming $x$ with $g^{-1}(a^2_{t_2})$ (i.e. zooming in instead of scaling down). We note that RNPs can be trained one depth at a time to decrease training time and resources.

To allow probabilistic sampling of programs, we can express an RNP as a structured variational auto-encoder \cite{kingma_auto-encoding_2014} (VAE) to learn an approximate posterior $q(z^K|x) \approx p(z^K|x) $ of an image $x$ given prior $p(z^K) \sim \mathcal{N}(0, 1)$, where $z^K$ is the highest level state vector. We therefore use an encoder network to parameterize the approximate posterior $q(z^K|x)$ and regularize \cref{eq-1} with the $KL(q||p)$ term.

\begin{figure}[!htbp]
    \centering
\includegraphics[width=\textwidth]{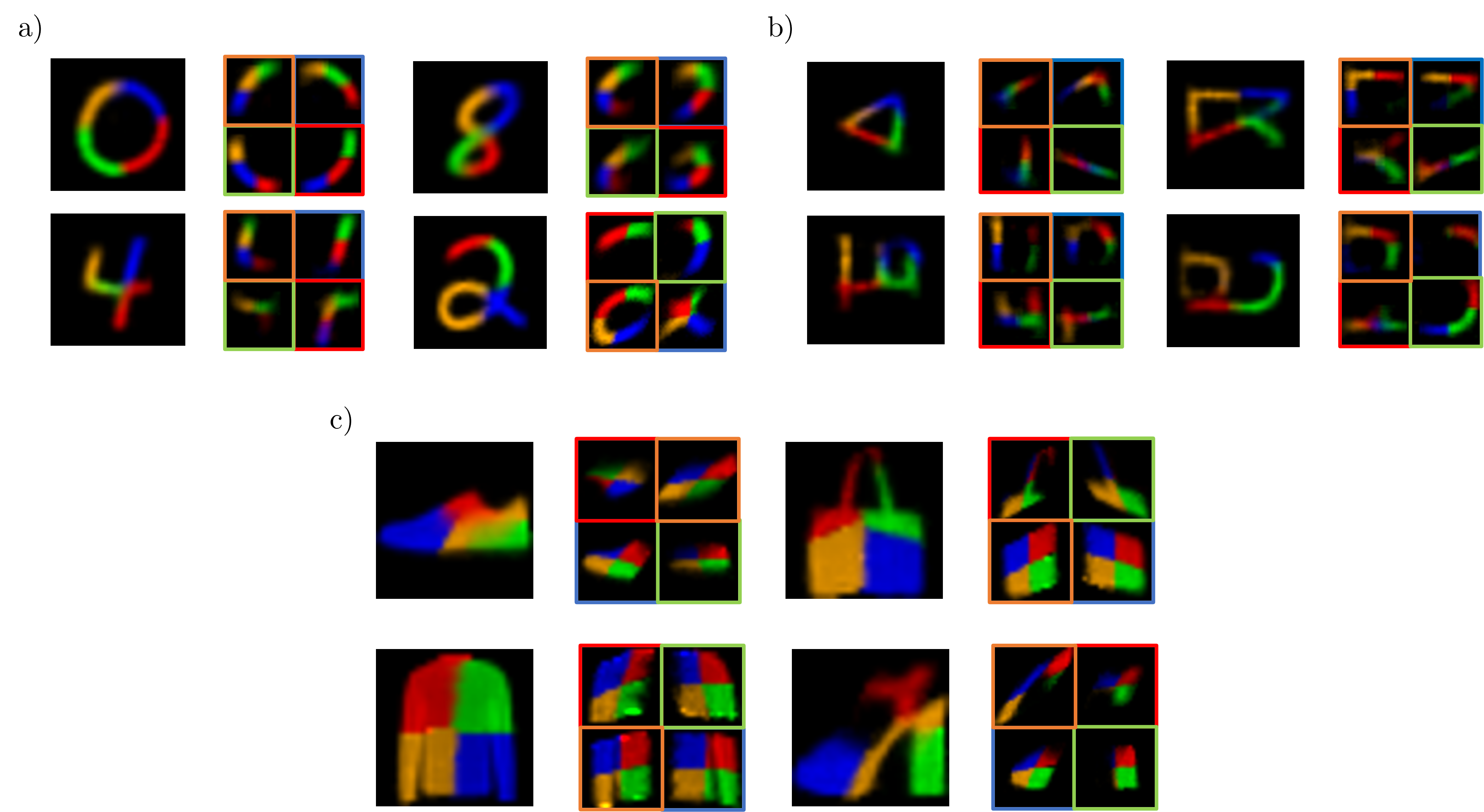}
  \caption{
    {\bf Hierarchical Parts-Based Decomposition by the Model after Learning:} Parsing of ({\bf a}) four MNIST digits, ({\bf b}) four Omniglot characters and ({\bf c}) four Fashion-MNIST objects by the model. Two levels of the hierarchical representation are shown, parts generated by $z^2$ (left; each part is denoted by a different color) and sub-parts generated by $z^1$ (right, bordered boxes, each sub-part is denoted by a different color). Order: blue $\rightarrow$ red $\rightarrow$ green $\rightarrow$ orange. Each bordered box shows the output of a program  generated by $z^1$ to construct a part as a combination of sub-parts transformed and placed within the reference frame of the part. 
  }
  \label{fig-3}

\end{figure}

\section{Results}

We first demonstrate how our RNPs can recursively  parse input images of MNIST digits \cite{lecun_et_al_yann_gradient-based_1998}, Omniglot characters \cite{lake_human-level_2015} and Fashion-MNIST objects \cite{xiao_fashion-mnist_2017} into parts and sub-parts. We then characterize the embedding space of state vectors at two levels and show how learned representations at various tree-depths can be composed to generate previously unseen image types.


\begin{figure}[h]
\centering
    \includegraphics[width=1\linewidth]{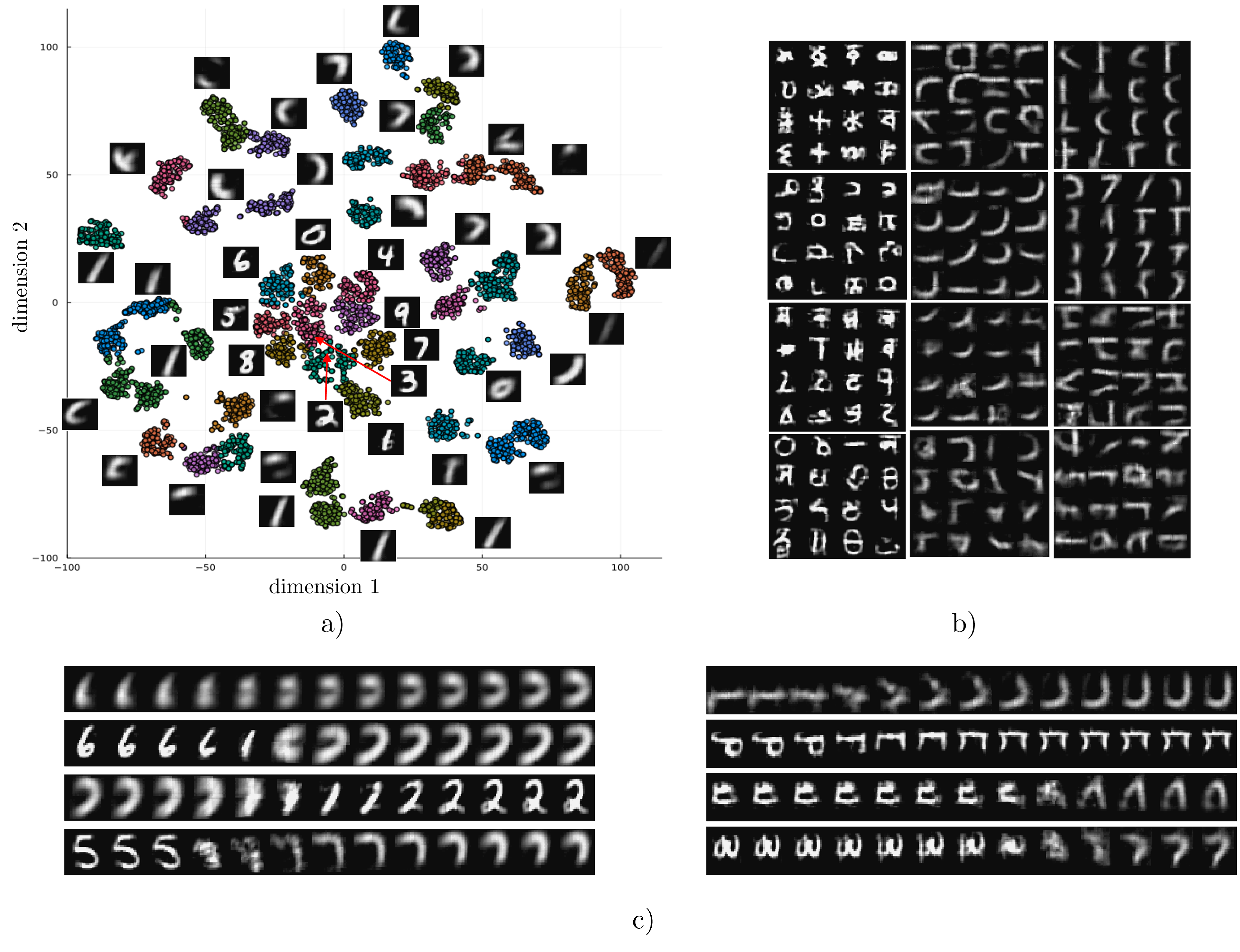}    
    \caption{
        \textbf{Topography of Neural Program Space.} ({\bf a}) t-SNE clustering of $z^2$ and $z^1$ vectors in a model trained on MNIST. A representative image is shown for each cluster. ({\bf b}) Example clusters of sampled images from $z^2$ (left column) and $z^1$ (remaining columns) for a model trained on Omniglot. ({\bf c}) Example linear interpolations in $z$ space from the center of one cluster (leftmost image) to the center of another cluster (rightmost image) show novel generated images from neural programs in the intermediate space. Left: MNIST, right: Omniglot.
        }
    \label{fig-4}
\end{figure}

\subsection{Image parsing into parts and sub-parts}

We trained RNPs to reconstruct MNIST digits and Omniglot characters as two-level generative programs. An encoder network 
was trained to map the input image to the top-level program (embedding vector) $z^2$. As described above, $z^2$ parameterizes $f^1_{state}$ and $f^1_{policy}$ via a  hypernetwork $\mathcal{H}$, and $z^1$ is the latent code corresponding to the parts (larger patches, 6x6px - 12x12px; \cref{fig-1}b). $z^1$ is then passed through the same hypernetwork $\mathcal{H}$ to synthesize sub-parts (smaller patches, 1.5x1.5px - 4x4px; \cref{fig-2}b). We force the network to learn a part-wise representation by constraining each part to be smaller than its parent, therefore requiring a sequence of steps to reconstruct it. Figure 3 shows examples of MNIST digits (\cref{fig-3}a), Omniglot characters (\cref{fig-3}b) and Fashion-MNIST objects (\cref{fig-3}c) generated by RNPs, with reconstructions at the level of parts (untiled-) and sub-parts (tiled images).

\subsection{Topography of neural programs}

A notable challenge in optimizing and representing probabilistic programs has been the absence of a continuous program space that can be interpretably manipulated.  As we use the same hypernetwork $\mathcal{H}$ to generate programs at all levels, we should expect that programs at different tree depths inhabit different areas of $|z|$-dimensional space, i.e. programs representing digits cluster separately from programs representing parts. Analyzing the embedding space of $z^2$ and $z^1$ vectors that represent the trained data (MNIST digits or Omniglot characters) reveals that $z^2$ and $z^1$ ``neural program'' vectors do cluster separately (\cref{fig-4}a,b). 

\begin{figure}[h]
\centering
    \includegraphics[width=\linewidth]{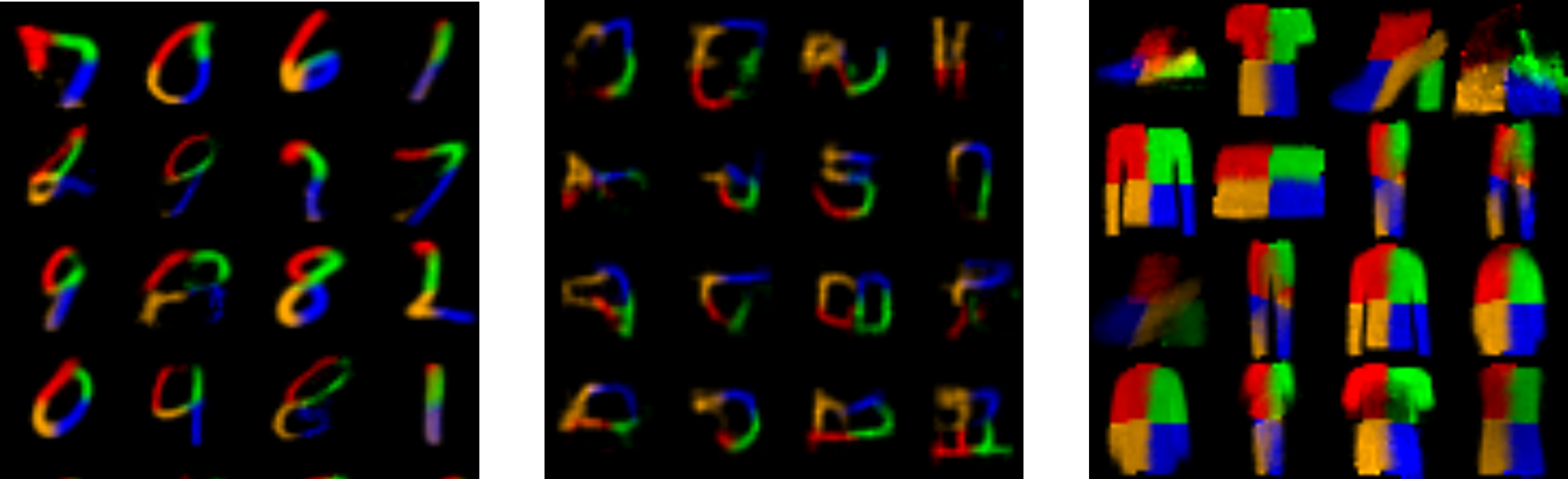}
    \caption{
        \textbf{Sampling from the Prior}. Sampling of $z^2$ from $\mathcal{N}(0, 1)$ for a model trained on MNIST digits (left), Omniglot characters (middle) and Fashion MNIST objects (right). As in (\cref{fig-3}), part order is blue $\rightarrow$ red $\rightarrow$ green $\rightarrow$ orange.
        }
    \label{fig-5}
\end{figure}

To test the expressiveness of our model, we investigated the space between learned $z^2$ and $z^1$ program clusters by linearly interpolating in the latent ``neural program" space occupied by the $z^2$ and $z^1$ vectors. Sampling from regions between clusters produced programs that generated novel images (\cref{fig-5}), showing that the model can exploit the latent structure of the program embedding space to synthesize previously unseen patterns by combining the learned parts.

\subsection{Compositionality and transfer learning}

Compositionality is a main goal of our architecture. With a generative model over programs, we are able to sample program space in regions outside those representing the trained data (\cref{fig-4}). This can be demonstrated by interpolating between clusters in $\{z^2, z^1\}$ (\cref{fig-4}c), or sampling randomly from $z^2 \sim \mathcal{N}(0, I)$ (\cref{fig-5}). Figure 5 shows that the model can generate novel characters by synthesizing learned primitives in different, often novel, combinations of parts. 

\begin{figure}[h]
\centering
    \includegraphics[width=\linewidth]{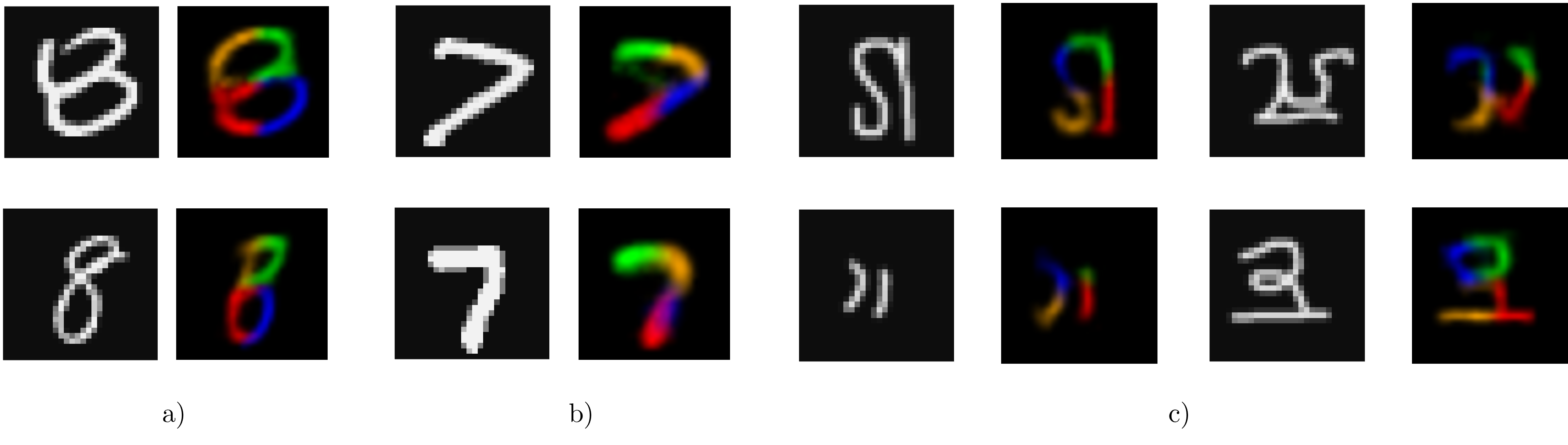}
    \caption{
        \textbf{Transfer Learning.} RNPs trained on a training set of classes (see text) are able to explain novel examples from unseen classes and synthesize the parts for MNIST digits ({\bf a,b}) and Omniglot characters ({\bf c}).
        }
    \label{fig-6}
\end{figure}

We further tested the compositional ability of our model in a transfer learning task. We trained RNPs on all MNIST classes but one ($7$ or $8$) and on the Omniglot transfer dataset. By adjusting the weights of the encoder network (but not the decoder hypernetwork $\mathcal{H}$), RNPs were able to synthesize parts for the unseen class (\cref{fig-6}).

\section{Conclusion}

In this paper, we introduced Recursive Neural Programs (RNPs), a new model for  differentiably learning tree-structured data as  sensory-motor sequences in a way that allows flexible composition of learned primitives using a recursive ``grammar.'' We demonstrated our model's ability to generate images using a hierarchy of parts and their transformations. Our architecture can also be applied to learning in arbitrary domains, such as audio, video and other dynamical processes such as motor behavior. 

There are several potential directions for future research. Using the same hypernetwork at different levels allows natural recursion, but limits the expressive power of the model. This can be addressed by learning different hypernetworks for different levels. Hypernetworks describing different data modalities (e.g. audio, visual, etc.) could be combined to generate richer multi-modal neural programs, provided constraints on the size of the primary network are taken into account  \cite{galanti_modularity_2020}. Training deep RNPs across levels and across time steps can be challenging. This could be addressed by training RNPs at different depths in parallel. Another potential area for improvement is replacing bilinear interpolation used for transformation of image primitives, which can result in poor quality gradients, with smoother functions to sample images (e.g. \cite{klocek_et_al_hypernetwork_2019}). Finally, message passing between nodes at different tree depths could allow for bidirectional information flow: predictions from parents to children, and belief updates from children to parents (using, e.g., prediction errors). We intend to explore such predictive coding-based architectures for RNPs in future work.

\section*{Acknowledgments}
We thank Dimitrios Gklezakos for his help with hypernetworks, and Preston Jiang for feedback on probabilistic aspects of the model. This material is based upon work supported by the Defense Advanced Research Projects Agency (DARPA) under Contract No. HR001120C0021, a Weill Neurohub Investigator award, a grant from the Templeton World Charity Foundation, and a Cherng Jia \& Elizabeth Yun Hwang Professorship to RPNR. The opinions expressed in this publication are those of the authors and do not necessarily reflect the views of the funders.

\medskip
\printbibliography

@misc{jiang_preston_l_dynamic_2021,
	title = {Dynamic {Predictive} {Coding} with {Hypernetworks} {\textbar} {bioRxiv}},
	url = {https://www.biorxiv.org/content/10.1101/2021.02.22.432194v2.abstract},
	author = {{Jiang, Preston L} and {Gklezakos, Dimitrios} and Rao, Rajesh P. N.},
	year = {2021},
}

@techreport{innes_dont_2018,
	title = {Don't {Unroll} {Adjoint}: {Differentiating} {SSA}-{Form} {Programs}},
	shorttitle = {Don't {Unroll} {Adjoint}},
	url = {https://ui.adsabs.harvard.edu/abs/2018arXiv181007951I},
	author = {Innes, Michael},
	month = oct,
	year = {2018},
	note = {Publication Title: arXiv e-prints
ADS Bibcode: 2018arXiv181007951I
Type: article},
}

@techreport{kingma_adam_2017,
	title = {Adam: {A} {Method} for {Stochastic} {Optimization}},
	shorttitle = {Adam},
	url = {http://arxiv.org/abs/1412.6980},
	number = {arXiv:1412.6980},
	institution = {arXiv},
	author = {Kingma, Diederik P. and Ba, Jimmy},
	month = jan,
	year = {2017},
	doi = {10.48550/arXiv.1412.6980},
	note = {arXiv:1412.6980 [cs]
type: article},
}

@techreport{innes_fashionable_2018,
	title = {Fashionable {Modelling} with {Flux}},
	url = {http://arxiv.org/abs/1811.01457},
	number = {arXiv:1811.01457},
	institution = {arXiv},
	author = {Innes, Michael and Saba, Elliot and Fischer, Keno and Gandhi, Dhairya and Rudilosso, Marco Concetto and Joy, Neethu Mariya and Karmali, Tejan and Pal, Avik and Shah, Viral},
	month = nov,
	year = {2018},
	doi = {10.48550/arXiv.1811.01457},
	note = {arXiv:1811.01457 [cs]
type: article},
}

@inproceedings{he_deep_2016,
	title = {Deep {Residual} {Learning} for {Image} {Recognition}},
	url = {https://openaccess.thecvf.com/content_cvpr_2016/html/He_Deep_Residual_Learning_CVPR_2016_paper.html},
	author = {He, Kaiming and Zhang, Xiangyu and Ren, Shaoqing and Sun, Jian},
	year = {2016},
	pages = {770--778},
}

@inproceedings{jaderberg_et_al_spatial_2015,
	title = {Spatial {Transformer} {Networks}},
	url = {https://proceedings.neurips.cc/paper/2015/hash/33ceb07bf4eeb3da587e268d663aba1a-Abstract.html},
	booktitle = {Advances in {Neural} {Information} {Processing} {Systems}},
	author = {Jaderberg et al., Max},
	year = {2015},
}

@inproceedings{mnih_et_al_recurrent_2014,
	title = {Recurrent {Models} of {Visual} {Attention}},
	url = {https://proceedings.neurips.cc/paper/2014/hash/09c6c3783b4a70054da74f2538ed47c6-Abstract.html},
	booktitle = {Advances in {Neural} {Information} {Processing} {Systems}},
	author = {Mnih et al., Volodymyr},
	year = {2014},
}

@article{kaelbling_planning_1998,
	title = {Planning and acting in partially observable stochastic domains},
	volume = {101},
	issn = {00043702},
	url = {https://linkinghub.elsevier.com/retrieve/pii/S000437029800023X},
	doi = {10.1016/S0004-3702(98)00023-X},
	language = {en},
	number = {1-2},
	journal = {Artificial Intelligence},
	author = {Kaelbling, Leslie Pack and Littman, Michael L. and Cassandra, Anthony R.},
	month = may,
	year = {1998},
	pages = {99--134},
}

@inproceedings{socher_et_al_parsing_2011,
	title = {Parsing {Natural} {Scenes} and {Natural} {Language} with {Recursive} {Neural} {Networks}},
	url = {https://openreview.net/forum?id=SyEeunWObH},
	language = {en},
	author = {Socher et al., Richard},
	month = jan,
	year = {2011},
}

@article{galanti_modularity_2020,
	title = {On the {Modularity} of {Hypernetworks}},
	url = {http://arxiv.org/abs/2002.10006},
	journal = {arXiv:2002.10006 [cs, stat]},
	author = {Galanti, Tomer and Wolf, Lior},
	month = nov,
	year = {2020},
	note = {arXiv: 2002.10006},
}

@article{ha_hypernetworks_2016,
	title = {{HyperNetworks}},
	url = {http://arxiv.org/abs/1609.09106},
	journal = {arXiv:1609.09106 [cs]},
	author = {Ha, David and Dai, Andrew and Le, Quoc V.},
	month = dec,
	year = {2016},
	note = {arXiv: 1609.09106},
}

@article{lake_human-level_2015,
	title = {Human-level concept learning through probabilistic program induction},
	volume = {350},
	url = {https://www.science.org/doi/full/10.1126/science.aab3050},
	doi = {10.1126/science.aab3050},
	number = {6266},
	journal = {Science},
	author = {Lake, Brenden M. and Salakhutdinov, Ruslan and Tenenbaum, Joshua B.},
	month = dec,
	year = {2015},
	pages = {1332--1338},
}

@article{hinton_how_2021,
	title = {How to represent part-whole hierarchies in a neural network},
	url = {https://arxiv.org/abs/2102.12627v1},
	doi = {10.48550/arXiv.2102.12627},
	language = {en},
	journal = {arXiv:2102.12627},
	author = {Hinton, Geoffrey},
	month = feb,
	year = {2021},
}

@misc{lecun_et_al_yann_gradient-based_1998,
	title = {Gradient-based learning applied to document recognition {\textbar} {IEEE} {Journals} \& {Magazine} {\textbar} {IEEE} {Xplore}},
	url = {https://ieeexplore.ieee.org/document/726791},
	author = {{LeCun et al., Yann}},
	year = {1998},
}

@article{kingma_auto-encoding_2014,
	title = {Auto-{Encoding} {Variational} {Bayes}},
	url = {http://arxiv.org/abs/1312.6114},
	journal = {arXiv:1312.6114 [cs, stat]},
	author = {Kingma, Diederik P. and Welling, Max},
	month = may,
	year = {2014},
	note = {arXiv: 1312.6114},
}

@inproceedings{eslami_et_al_attend_2016,
	title = {Attend, {Infer}, {Repeat}: {Fast} {Scene} {Understanding} with {Generative} {Models}},
	shorttitle = {Attend, {Infer}, {Repeat}},
	url = {https://proceedings.neurips.cc/paper/2016/hash/52947e0ade57a09e4a1386d08f17b656-Abstract.html},
	booktitle = {Advances in {Neural} {Information} {Processing} {Systems}},
	author = {Eslami et al., S. M. Ali},
	year = {2016},
}

@techreport{gklezakos_active_2022,
	type = {preprint},
	title = {Active {Predictive} {Coding} {Networks}: {A} {Neural} {Solution} to the {Problem} of {Learning} {Reference} {Frames} and {Part}-{Whole} {Hierarchies}},
	shorttitle = {Active {Predictive} {Coding} {Networks}},
	language = {en},
	institution = {Neuroscience},
	author = {Gklezakos, Dimitrios C. and Rao, Rajesh P. N.},
	month = jan,
	year = {2022},
	doi = {10.1101/2022.01.20.477125},
}

@techreport{xiao_fashion-mnist_2017,
	title = {Fashion-{MNIST}: a {Novel} {Image} {Dataset} for {Benchmarking} {Machine} {Learning} {Algorithms}},
	shorttitle = {Fashion-{MNIST}},
	url = {http://arxiv.org/abs/1708.07747},
	number = {arXiv:1708.07747},
	institution = {arXiv},
	author = {Xiao, Han and Rasul, Kashif and Vollgraf, Roland},
	month = sep,
	year = {2017},
	doi = {10.48550/arXiv.1708.07747},
	note = {arXiv:1708.07747 [cs, stat]
type: article},
}

@article{hinton_matrix_2018,
	title = {Matrix {Capsules} with {EM} {Routing}},
	language = {en},
	journal = {ICLR},
	author = {Hinton, Geoffrey and Sabour, Sara and Frosst, Nicholas},
	year = {2018},
	pages = {15},
}

@inproceedings{klocek_et_al_hypernetwork_2019,
	title = {Hypernetwork {Functional} {Image} {Representation}},
	isbn = {978-3-030-30493-5},
	doi = {10.1007/978-3-030-30493-5_48},
	language = {en},
	booktitle = {Artificial {Neural} {Networks} and {Machine} {Learning} – {ICANN} 2019: {Workshop} and {Special} {Sessions}},
	author = {Klocek et al., Sylwester},
	year = {2019},
}

\section*{Supplementary Material}

\subsection*{Model algorithm}

\begin{algorithm}[H]
    \caption{Two-level image generation}\label{alg1}
    \begin{algorithmic}[1]
        \State $z^2 \gets Encoder(x)$
        \State $enc^1, f^1_{state}, f^1_{policy}, dec^1_{state}, dec^1_{policy} \gets H(z^2)$
        \State $a^1_0, z^1_0 \gets init(z^2)$
        
        \For{$t_2 = 1:\tau_2$}
        \State $z^1_{t_2} = f^2_{state}(a^1_{t_2 -1}, z^1_{t_2 -1})$
        \State $a^1_{t_2} = f^2_{policy}(a^1_{t_2 -1}, z^1_{t_2 -1})$        
        \State $\hat{x}^2 = dec^1_{state}(z^1_{t_2})$
        \LeftComment{Bottom-level loop}
        \State $enc^0, f^0_{state}, f^0_{policy}, dec^0_{state}, dec^0_{policy} \gets H(z^1)$

        \State $a^0_0, z^0_0 \gets init(z^1)$

        \For{$t_1 = 1:\tau_1$}
        \State $z^0_{t_1} = f^1_{state}(a^0_{t_1 -1}, z^0_{t_1 -1})$
        \State $a^0_{t_1} = f^1_{policy}(a^0_{t_1 -1}, z^0_{t_1 -1})$
        \State $\hat{x}^1_t = dec^0_{state}(z^0_{t_1})$

        \State $p^1_t \gets p^1 + g(\hat{x}^1_t, a^0_t)$ \Comment{transformed reconstruction}

        \EndFor
        \State $p^2_t \gets p^2 + g(p^1, a^1_t)$
        \EndFor

    \end{algorithmic}
\end{algorithm}

\subsection*{Model and training details}

\subsubsection*{Parameters}

RNPs consist of a hypernet $\mathcal{H}$ that generates the parameters of an autoregressive network for a level $k$. All hypernetworks used in this study are 6-layer MLP's (64 units, elu activations), with seven heads, parameterizing networks of the same size (except to reflect different sizes of $|z|$). All networks for a given $k$ consisted of fully connected layers of 64 units, except the RNNs $f^k_{state}$ and $f^k_{policy}$, which retained the dimensionality of $|z|$. 

The encoder network consisted of five ResNet blocks (32 channels) \cite{he_deep_2016} and four fully connected layers (64 units).

\subsubsection*{Training}
We trained all models using the ADAM optimizer \cite{kingma_adam_2017} with a learning rate of 4e-5, which reliably showed convergence. We trained our models for 200 epochs, except on the Omniglot dataset where we trained for 400 epochs. We used $|z| = 32$ on MNIST and Fashion-MNIST, and $|z| = 96$ for Omniglot. Models were trained on a single GPU (Nvidia Quadro RTX 6000).

\subsubsection{Software}

All experiment code was written in Julia using Flux.jl \cite{innes_fashionable_2018} and Zygote.jl \cite{innes_dont_2018}. Code is publicly available at \url{https://github.com/FishAres/RNP}.

\end{document}